\documentclass[letterpaper, 10 pt, conference]{ieeeconf}  % Comment this line out if you need a4paper

\IEEEoverridecommandlockouts                              % This command is only needed if 
\usepackage{amsfonts}
\usepackage{amsmath}
\usepackage{bbold}
\usepackage{algorithmic}
\usepackage{graphicx}
\usepackage{textcomp}
\usepackage{xcolor}
\usepackage[final]{changes}  %show changes
\usepackage{pifont}
\usepackage{multirow}
\usepackage{booktabs}
\usepackage{pgfplots}
\usepackage{subcaption}
\usepackage{xspace}
\usepackage{tikz}
\pgfplotsset{compat=1.18}
\usepgfplotslibrary{colorbrewer}
\newcommand\doubleplus{+\kern-1.3ex+\kern0.8ex}
\newtheorem{definition}{Definition}

\newcommand{\modelname}[0]{\textsc{PedCA-FT}\xspace}

\title{Early Risk Prediction of Pediatric Cardiac Arrest from Electronic Health Records via Multimodal Fused Transformer}

\author{Jiaying Lu\textsuperscript{*}$^{1}$\textsuperscript{\ding{41}}, Stephanie R. Brown\textsuperscript{*}$^{2}$, Songyuan Liu$^{1}$, Shifan Zhao$^{3}$, Kejun Dong$^{1}$, Del Bold$^{1}$,\\ 
 Michael Fundora$^{2}$, Alaa Aljiffry$^{2}$, Alex Fedorov$^{1}$, Jocelyn Grunwell$^{2}$ and Xiao Hu$^{1,4}$% <-this % stops a space
\thanks{\textsuperscript{\ding{41}}Corresponding Author: Jiaying Lu, Ph.D. (jiaying.lu@emory.edu)}% 
\thanks{\textsuperscript{*}These authors contributed equally to this work}
\thanks{$^{1}$Center for Data Science, Nell Hodgson Woodruff School of Nursing, Emory University, Atlanta, GA 30322, USA}%
\thanks{$^{2}$Department of Pediatrics, Children's Healthcare of Atlanta Cardiology, School of Medicine, Emory University, Atlanta, GA 30322, USA}%
\thanks{$^{3}$Department of Mathematics, Laney Graduate School, Emory University, Atlanta, GA 30322, USA}%
\thanks{$^{4}$The Wallace H. Coulter Department of Biomedical Engineering, Georgia Institute of Technology \& Emory University, Atlanta, GA 30322, USA}%
}

\begin{document}

\maketitle
\thispagestyle{empty}
\pagestyle{empty}

%%%%%%%%%%%%%%%%%%%%%%%%%%%%%%%%%%%%%%%%%%%%%%%%%%%%%%%%%%%%%%%%%%%%%%%%%%%%%%%%
%7 pages w/ references. Reviews are traditional single blind peer reviews where the identity of the reviewer is not visible to the author(s).
\begin{abstract}
Early prediction of pediatric cardiac arrest (CA) is critical for timely intervention in high-risk intensive care settings. We introduce \modelname, a novel transformer-based framework that fuses tabular view of EHR with the derived textual view of EHR to fully unleash the interactions of high-dimensional risk factors and their dynamics. By employing dedicated transformer modules for each modality view, \modelname captures complex temporal and contextual patterns to produce robust CA risk estimates. Evaluated on a curated pediatric cohort from the CHOA-CICU database, our approach outperforms ten other artificial intelligence models across five key performance metrics and identifies clinically meaningful risk factors. These findings underscore the potential of multimodal fusion techniques to enhance early CA detection and improve patient care.
\end{abstract}

%%%%%%%%%%%%%%%%%%%%%%%%%%%%%%%%%%%%%%%%%%%%%%%%%%%%%%%%%%%%%%%%%%%%%%%%%%%%%%%%
\renewcommand*{\thefootnote}{\fnsymbol{footnote}}

% main sections
\section{Introduction}

%para1: Cardiac Arrest in Pediatric Population (CICU)
Cardiac arrest (CA) in the pediatric population is a critical and life-threatening event with significant implications for morbidity and mortality~\cite{frazier2021risk}. While less common than in adults, pediatric CA is often associated with a preceding period of physiological instability, making timely identification and intervention crucial. Many cases of CA in children occur in intensive care units (ICUs)~\cite{zeng2020pic}, particularly cardiac ICUs~\cite{morgan2021cicu}, where children with congenital or acquired heart diseases are managed. In these high-stakes environments, the interplay of complex cardiac conditions, surgical interventions, and critical care therapies, coupled with significant cognitive and data overload, creates a unique set of challenges for clinicians~\cite{winters2018society}. 
\added{Early CA risk prediction refers to the use of clinical data available within the first few hours of admission~\cite{moore2025prognostic,brown2025MLCA} to identify patients at risk of experiencing CA later in their hospitalization.}
Such early identification enables proactive measures to prevent arrests and provide situational awareness. This highlights the importance of developing robust predictive models tailored to the pediatric cardiac ICU population, integrating clinical, physiological, and potentially novel data sources.

%para2: challenges of EHR based CA risk prediction. 1. heterogeneous data modalities; 2. multi-resolution.
Electronic Health Records (EHRs) offer a rich repository of patient data~\cite{zeng2020pic}, making them an attractive resource for cardiac arrest risk prediction. By aggregating diverse information such as demographics, vital signs, laboratory results, and clinical notes, EHRs provide a comprehensive view of a patient’s health trajectory. However, leveraging EHR data for pediatric CA risk prediction is not without challenges~\cite{wornow2023ehrshot}. 
First, EHRs are inherently \textit{heterogeneous}\footnote[2]{Heterogeneous: while many works use the term “multivariate” to refer to the complex input features in the context of time-series models, here we use “heterogeneous” to emphasize two levels of heterogeneity: (1) static versus dynamic risk factors, and (2) categorical, numerical, and textual input formats.}, encompassing various modalities that differ in format and clinical significance. For examples, demographics are stored in the static categorical format that serve as underlying risk factors, whereas vital signs are recorded as longitudinal numerical measurements that reflect short term physiological status.
Second, EHR data are \textit{multi-resolution}: some measurements, like vital signs, are updated in near real-time, while others, such as lab results, are recorded at more sparse intervals. 
Traditionally, clinicians have relied on manually developed criteria and scoring systems (\textit{e.g.}, PEWS~\cite{monaghan2005pews}, IDO2~\cite{loomba2023inadequate}) to predict cardiac arrest. While these systems are valuable, they may not fully capture the subtle, individual-specific nuances of a patient’s condition because they depend on universally predefined thresholds applied to a fixed set of clinical observations. Moreover, certain systems, such as IDO2, are proprietary, limiting their accessibility without incurring additional costs.
In contrast, data-driven approaches leverage EHR data through advanced artificial intelligence (AI) techniques, leading to improved performance in early detection and risk stratification of pediatric cardiac arrest~\cite{brown2025MLCA}.

% para 3: gap of existing ML. Tabular model can handle heterogenity, but not multi-resolution. Time-series model can handle multi-resolution to some degree, but not much. Our solution is use fusion transformer.
The heterogeneous and multi-resolution nature of EHR poses significant challenges to AI models for EHR-based pediatric CA risk prediction. 
Conventional AI models (e.g., random forest, XGBoost~\cite{chen2016xgboost}), which are designed for static tabular data, excel at handling high-dimensional numerical and categorical risk factors by capturing subtle and non-linear patterns indicative of impending cardiac arrest. However, these models operate on fixed-dimensional input features and typically require ad-hoc, domain-specific feature preprocessing—such as aggregation~\cite{rajkomar2018countbase}—which can overlook important dynamic temporal changes in risk factors. 
On the other hand, time-series AI models~\cite{goswami2024momentfamilyopentimeseries} are well-suited for analyzing longitudinal data, as they can capture temporal dependencies and trends. Although resolution unification techniques (\textit{e.g.} resampling, padding, interpolation)~\cite{rasul2023lag,ekambaram2024tiny} can address the multi-resolution challenges inherent in EHR data, time-series models often struggle to effectively handle high-dimensional inputs.

% para4: Introduce our model that address both challenges.
To address the above mentioned challenges, we propose our innovative approach, namely \textbf{PED}iatric \textbf{C}ardiac \textbf{A}rrest prediction via \textbf{F}used \textbf{T}ransformer (\modelname), based on a novel tabular-textual multimodal fusion strategy~\cite{shi2021multimodal,lu2023mug}, and the powerful Transformer backbone model for each modality.
We view the heterogeneous and multi-resolution risk factors as two complementary modalities, tabular and textual, each characterized by its own intrinsic data structure. Specifically, we employ a dedicated tabular Transformer~\cite{huang2020tabtransformer,gorishniy2021FT-Trans} to effectively handle high-dimensional static and aggregated longitudinal tabular features, while a pre-trained textual EHR Transformer~\cite{yang2022gatortron} processes the textual representations~\cite{contreras2024dellirium} derived from the original EHR data. A fusion Transformer is then used to integrate the modality-specific representations, ultimately computing the probability of CA onset risk.
We evaluate the effectiveness of our \modelname on a curated cohort from the CHOA-CICU database comprising 3,566 pediatric patients with a $4.0\%$ incidence of CA. Our model, along with ten other AI models, is compared using 5-fold cross-validation. Through this comprehensive evaluation, our proposed approach marginally outperforms all compared models. Furthermore, a feature importance analysis reveals that several of the identified risk factors align well with clinical knowledge for pediatric CA, as confirmed by our clinical collaborators.

\section{Study Design}
\label{sec:study}
\subsection{Problem Definition}
We are interested in the early risk prediction of cardiac arrest for pediatric ICU population.  
\begin{definition}[Patient EHR Data]
The EHR data $\mathbf{x}$ of one patient include static risk factors $\mathbf{x}^{(s)} \in \mathbb{R}^{d_s}$ during one admission such as demographics and admission diagnosis, and temporal risk factors $\mathbf{x}^{(t)}_{0:T} \in \mathbb{R}^{d_t\times T}$ such as patient's lab results, vital signs, and nursing assessments. Therefore, the EHR data within first $\tau$ hours is defined by $\mathbf{x}_{{t\leq \tau}}=(\mathbf{x}^{(s)}, \mathbf{x}^{(t)}_{0:\tau})$, and $|\mathbf{x}|$ denotes overall number of risk factors contained in EHR data. 
\end{definition}

\begin{definition}[Early Risk Prediction of Cardiac Arrest]
We aim to employ a predictive model $f_\theta(\cdot)$ using a patient EHR data $\mathbf{x}_{t\leq \tau}$ available in the first $\tau$ (\added{$\tau=24$ in our experiments}) hours of admission, to predict the probability of developing cardiac arrest $\hat{y}=f_\theta(\mathbf{x}_{t\leq \tau})$, where $\hat{y} \in \{0, 1\}$ during the rest of the admission stay \added{(\textit{e.g.} CA onset can occur up to several weeks after admission)}.
\end{definition}

\subsection{Data Source and Cohort Selection}
CHOA-CICU database is a private pediatric intensive care database extracted from electronic health record for all pediatric patients (age less than 18 years) admitted from 1/1/2018 to 12/31/2023 to a large, quaternary, academic Pediatric Cardiac Intensive Care Unit (CICU) at Children's Hospital at Atlanta and Emory University. 
We obtain 9 static risk factors from CHOA EHR, including: ``sex, ethnicity, first race, second race, primary language, respiration distress, age, admission diagnoses, admission ICD-9 codes''. Among 184 temporal risk factors, we obtain 129 vital signs and nursing assessments from flow sheet after removing non-relevant items and removing missing rate $> 0.5$ items. We also obtain 49 lab test results after removing missing rate $> 0.5$ items, and 5 medications after removing missing rate $> 0.7$ items.
Table~\ref{tab:data_stat} briefly presents the number of patients included in the two data sources. 
%We further conduct cohort selection to exclude certain patients.

\begin{table}[ht]
\centering
\caption{Statistics of CICU patients After Cohort Selection.}
%Dx denotes diagnosis, Vs denotes vital signs, Rx denotes prescriptions, Px denotes procedures, Sx denotes symptoms.}
\label{tab:data_stat}
\begin{tabular}{c|c}
\toprule
Category & Value\\
\midrule
Patient, no. (\% of case)& 3,566 (4.0\%) \\
Admissions, no. (\% of case) & 4,672 (3.1\%)\\
\hline
Age, median years (Q1--Q3) & 0.6 (0.1 -- 5.5)\\
Sex, no. of female (\%) & 1,676 (47.0\%)\\
CICU LoS, median days (Q1--Q3) & 8.4 (4.3 - 20.0)\\
HLoS, median days (Q1--Q3) & 2.0 (0.9 - 5.2)\\
Mortality, no., (\% of Admission) & 185 (4.0\%)\\
\hline
Static Risk Factors, no.& 9\\
Temporal Risk Factors, no. & 184\\
-(Tempo) Vital\&Assessments, no. & 129\\
-(Tempo) Lab Results, no. & 49\\
-(Tempo) Medications, no. & 5\\
\bottomrule
\end{tabular}
\vspace{-0.35cm}
\end{table}

% \begin{table*}[ht!]
% \centering
% \caption{Statistics of CA patients After Cohort Selection. Dx denotes diagnosis, Vs denotes vital signs, Rx denotes prescriptions, Px denotes procedures, Sx denotes symptoms. \jiaying{we want to change to a single column table. With a bit more patient characteristics: length of stay.}}
% %\label{tab:data_stat}
% \begin{tabular}{c|ccc|cc|p{3cm}}
% \toprule
% data source& CA Pat. (Adm.) & Ctrl Pat. (Adm.) & Incidence Pat.\% (Adm.\%) & Age-Days Mean (Q1–-Q3) & \%Female & Risk Factors \\
% \midrule
% %PIC & 358 (359) & 11,645 (12,168) & 3.0\% (2.9\%) &  916 (40 -- 1256)& 42.6\% & (static) Demogr, adm\_Dx; (ts) Vs, Lab, Rx, OR\_Px\\
% \hline
% %CHOA-CICU & 176 (177) & 7,484 (17,327) & 2.30\% (1.01\%) & & & (static) Demogr, adm\_Dx; (ts) Vs, Lab, Rx\\ 
% CHOA-CICU & 143 (146) & 3,423 (4,526) &  4.0\% (3.1\%) & 231 (36 -- 2,008) & 47.0\%& (static) Demogr, adm\_Dx; (ts) Vs, Lab, Med \\
% \bottomrule
% \end{tabular}
% \end{table*}

\section{Methods}
\begin{figure*}[ht!]
\centering
\includegraphics[width=\linewidth]{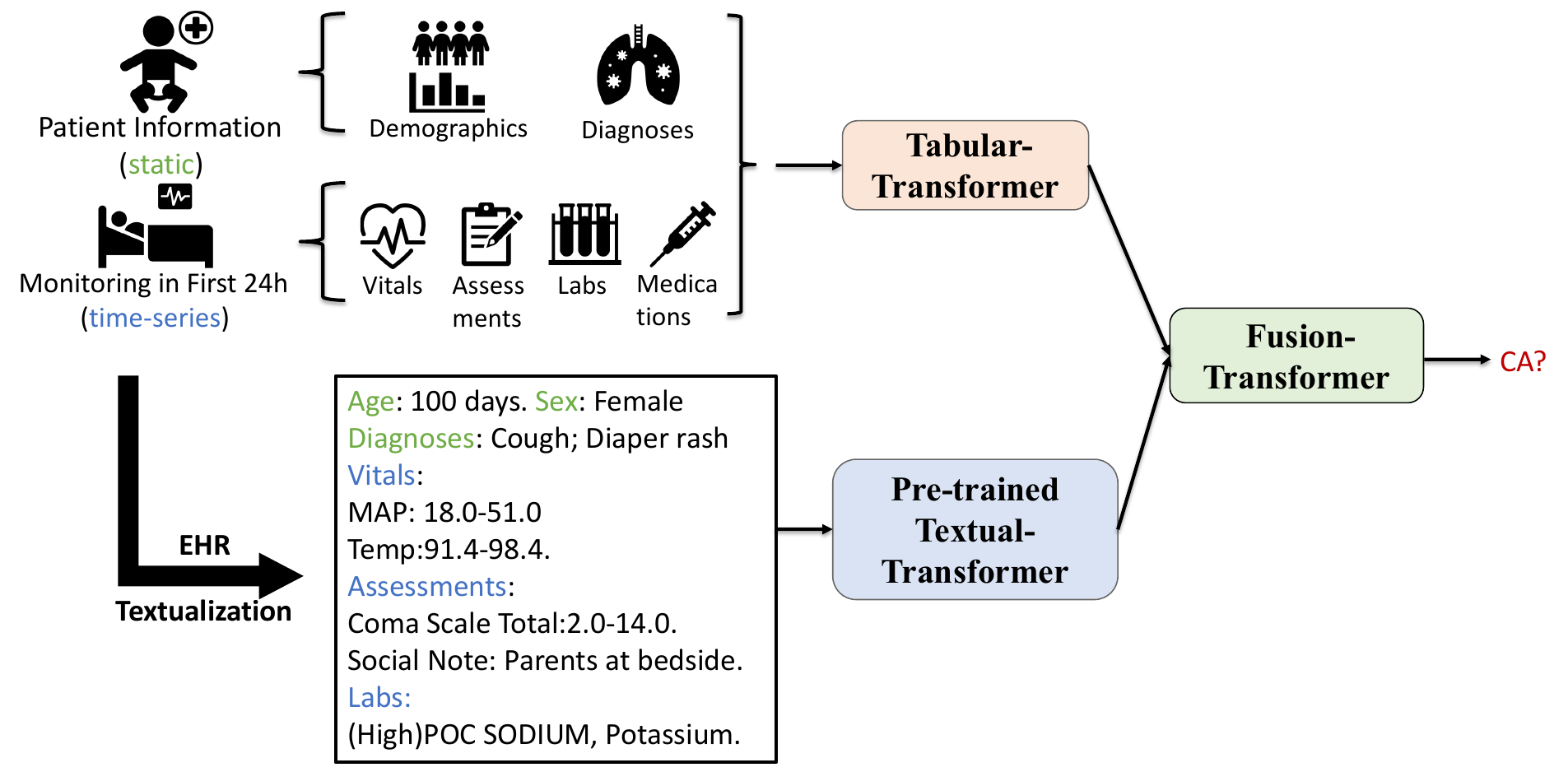}
\caption{Model Architecture of \modelname.}
\vspace{-0.4cm}
\label{fig:model}
\end{figure*}

The two major challenges we face are (1) the large feature space due to the multimodal, heterogeneous patient risk factors; (2) the label imbalance due to the low prevalence of CA in pediatric CICU patients.  
Our proposed model, \modelname, is a multimodal transformer late-fusion model, which utilize both raw structured EHR features (tabular) and derived textualized EHR features (textual) as input. As can be seen in Fig.~\ref{fig:model}, \modelname ~consists of three modules: tabular-transformer $\mathcal{T}_{tab}$, pre-trained textual-transformer $\mathcal{T}_{txt}$, and fusion-transformer $\mathcal{T}_{fus}$. The tabular-transformer is corresponding to obtain latent representation from patient's raw EHR data, denoted by $\mathbf{h}_{tab}=\mathcal{T}_{tab}(\mathbf{x}_{t\leq \tau})$. The textual-transformer is corresponding to obtain latent representation from patient's textualized EHR data, denoted by $\mathbf{h}_{txt}=\mathcal{T}_{txt}(g(\mathbf{x}_{t\leq \tau}))$ where $g(\cdot)$ denotes the textualization function for EHR. After obtaining latent representations from two views, we utilize the fusion-transformer to predict the likelihood of cardiac arrest, denoted by $\hat{y}=\mathcal{T}_{fus}(\mathbf{h}_{tab},\mathbf{h}_{txt})$.

\subsection{Tabular Transformer}
Tabular models (\textit{e.g.} XGBoost~\cite{chen2016xgboost}, LightGBM) have demonstrated excessive performance in EHR-based risk predictions. Here, we also employ a tabular view to handle the numerical and categorical risk factors that widely exist in the EHR. We follow the common practice to preprocess these tabular features, where numerical ones are passed through without change and categorical ones are encoded as monotonically increasing integers. 
For the static risk factors $\mathbf{x}^{(s)}$, we keep them as the original format. For the temporal risk factors, aggregation process is needed to accommodate the input format requirements. Therefore, 
\begin{equation}
    \mathbf{x}^{(t)}_{\text{agg}} = \phi(\mathbf{x}^{(t)}_{0:\tau}),\label{eq:agg}
\end{equation}
where $\phi: \mathbb{R}^{d_t \times \tau}\rightarrow \mathbb{R}^{d_t}$ denotes the aggregation operator that maps the time-series features into tabular features. A wide range of aggregation operation can be used for $\phi$, and we opt to $LAST(\cdot)$ which keeps the last observed element from the whole time-series. The input of tabular-transformer $\mathcal{T}_{tab}$ is $\mathbf{x}_{t\leq\tau}=(\mathbf{x}^{(s)},\mathbf{x}^{(t)}_{\text{agg}})$ after applying Eq.~\eqref{eq:agg}. 

We adapt a tabular feature oriented Transformer~\cite{huang2020tabtransformer,gorishniy2021FT-Trans} variation as our tabular view. In a nutshell, the numerical and categorical inputs are first transformed to dense embeddings and then feed into a stack of Transformer layers. The dense embedding for $i$-th factor $x_{(i)} \in \mathbb{X}_i$ of $\mathbf{x}_{t\leq\tau}$ is computed by
\begin{equation}
    \mathbf{e}_{(i)}=\xi_{(i)}(x_{(i)})+\mathbf{b}_{(i)},
\end{equation}
where $\mathbf{e}_{(i)}(\cdot) \in \mathbb{R}^d$ is $i$-th the transformed dense embedding, $\xi_{(i)}: \mathbb{X}_{(i)} \rightarrow \mathbb{R}^d$ is the $i$-th transformation function, and $\mathbf{b}_{(i)} \in \mathbb{R}^d$ is $i$-th feature transformation bias. 
Specifically, we implement $\xi_{(i)}(\cdot)$ as the element-wise multiplication 
\begin{equation}
\xi_{(i)}(x_{(i)}^{num})=x_{(i)}^{num}\cdot \mathbf{W}_{(i)}^{num}
\end{equation}
with the learnable vector $\mathbf{W}_{(i)}^{num} \in \mathbb{R}^d$ for numerical format features, and $\xi_{(i)}$ as the categorical embedding lookup 
\begin{equation}
\xi_{(i)}(x_{(i)}^{cat})=\mathbf{W}_{(i)}^{cat} \mathbb{1}(x_{(i)}^{cat})
\end{equation}
with one-hot encoding function $\mathbb{1}(\cdot): \mathbb{X}_{(i)}\rightarrow \mathbb{R}^d$ and the lookup table $\mathbf{W}_{(i)}^{(cat)} \in \mathbb{R}^{c_{i}\times d}$. 
Overall, 
\begin{equation}
    \mathbf{e}_{tab}= stack(\mathbf{e}_{(1)}^{num},\dots,\mathbf{e}_{(|\mathbf{x}^{num}|)}^{num},\mathbf{e}_{(1)}^{cat},\dots,\mathbf{e}_{(|\mathbf{x}^{cat}|)}^{cat}),
\end{equation}
where $\mathbf{e}_{tab} \in \mathbb{R}^{\mathbf{|x|}\times d}$. Then, L standard general Transformer layers $\mathcal{F}_1, \dots, \mathcal{F}_L$ are applied 
\begin{equation}
\mathbf{h}_k=\mathcal{F}_{k}(\mathbf{h}_{k-1})
\label{eq:multi_transformer}
\end{equation}
with $\mathbf{h}_{0}=\mathbf{e}_{tab}$ and $\mathbf{h}_L\in\mathbb{R}^{h_{tab}}$. Please refer to the original paper~\cite{vaswani2017attention} for technical details of the multi-head self-attention of transformer layer $\mathcal{F}$. 

\subsection{Textual Transformer}
The drawback of tabular view is that the aggregation process simplifies the time-series risk factors. We propose a textual view to fully capture the trend and change of these time-series risk factors by presenting their value ranges. Moreover, the static risk factors can also be effectively captured by directly presenting them in the textual format. A pre-trained textual-transformer $\mathcal{T}_{txt}$ is employed to obtain the latent embedding of the textualized EHR data $\mathbf{D}=g(\mathbf{x}_{t\leq \tau})$. 
In general, the textualized representation of $i$-th factor $x_{(i)} \in \mathbb{X}_i$ of $\mathbf{x}_{t\leq\tau}$ is computed by
\begin{equation}
\mathbf{D}_{(i)}=g_{(i)}(\mathbf{x}_{(i)}),
\end{equation}
where $g_{(i)}(\cdot)$ is the i-th textualization function, and $D_{(i)}$ denotes a sequence of length $|D_{(i)}|$ textual tokens $\{d_1,d_2,\dots,d_{|D_{(i)}|}\}$. 
Specifically, we implement $g_{(i)}(\cdot)$ using the following template for static feature $x_{(i)} \in \mathbb{X}_{(i)}$
\begin{equation}
g_{(i)}(x_{(i)})=\text{factorName}(x_{(i)})\doubleplus \text{``:"}\doubleplus x_{(i)}
\end{equation}
with $\text{factorName}(\cdot)$ denotes retrieving the factor name of $x_{(i)}$, and $\doubleplus$ denotes text concatenation function. For instance, $g_{(i)}(x_{(i)})=\text{``Sex: Female"}$ when $x_{(i)}=\text{``Female"}$ indicating the patient's sex.
Moreover, we implement $g_{(i)}(\cdot)$ using the following template for variable length time-series feature $\mathbf{x}_{(i),0:\tau}$  
\begin{multline}
g_{(i)}(\mathbf{x}_{(i),0:\tau})=\text{factorName}(\mathbf{x}_{(i)})\doubleplus \text{``:"} \doubleplus\\
\left\{
    \begin{array}{ c l }
    \textrm{set}(\mathbf{x}_{(i),0:\tau}), & \quad \textrm{if } \mathbf{x}_{(i)} \textrm{ is categorical}\\
    \min\text{--}\max(\mathbf{x}_{(i),0:\tau}), & \quad \textrm{otherwise},\\
    \end{array}
\right.\label{eq:ts_textual}
\end{multline}
where we keep all distinct values of each categorical time-series risk factor, and value min-max range of each numerical time-series factor. As an illustration, for categorical time-series risk factor \textit{skin color}, the textualized result is ``Skin Color: Pink; Pale.'' which covers all recorded values. Similarly, for numerical time-series risk factor \textit{body temperature}, the textualized result is ``Body Temp: 91.4--98.4.''. 
With the help of proposed textualization process, the heterogeneous multimodal EHR data is represented into a unified textual format which can be effectively handled by textual transformer, capturing dynamic trend of time-series. 

Pre-trained on massive textual corpus that covers public EHR databases, transformer-based models have shown promising in adults-centered risk prediction tasks~\cite{lu2024UQFM,chen2024adapting,shi2024ehragent}.
We employ a standard bidirectional encoder-only transformer (\textit{i.e.}, BERT~\cite{kenton2019bert}) for the pediatric cardiac arrest prediction, since it essentially is a sequence prediction problem after textualization of EHR. Follow the common practice, we add a special sequence token ``[CLS]'' at the beginning of the converted text, which is used as the text-view representation of one patient's EHR, 
\begin{equation}
\mathbf{h}_{BERT} = \text{BERT}(\text{[CLS]}, \mathbf{D}),
\end{equation}
where $\mathbf{h}_{BERT}\in \mathbb{R}^{(|\mathbf{D}|+1)\times d_{txt}}$ and $\text{BERT}(\cdot)$ denotes the employed bi-directional transformer model. It is worth noting that BERT here is different from the multi transformer layer in Eq.~\eqref{eq:multi_transformer}, since $\text{BERT}(\cdot)$ contains additional word embedding layer and its all learnable parameters are pre-trained. We use the ``[CLS]'' corresponding embedding as the text-view embedding of textualized EHR, thus $h_{txt}=h_{BERT}[0,:] \in \mathbb{R}^{d_{txt}}$.
% add special design for lab results: only keep abnormal.
One technical challenge for our textual transformer module is the maximum sequence length. To accommodate with that constraint, we further polish the textualization function $g_{}(\cdot)$ for time-series laboratory test results. Inspired by the clinical decision making process that paying more attentions to abnormal findings~\cite{jung2009clinical}, we filter out laboratory results that fall into the normal reference range and only keep the names of abnormal ones. Therefore, an example of textualized laboratory results can be ``(High) Sodium; Potassium. (Low) Glucose.''.
Furthermore, we add section headers, including ``Demographics'', ``Vitals'', ``Assessments'', ``Labs'' and ``Medications'',  for each covered EHR table, as can be seen in the toy example of Fig.~\ref{fig:model}. These above ad-hoc design aims to make the textualized EHR more readable and concise for the textual-transformer.

\subsection{Fusion Transformer}
After obtain the view-specific dense representation of patient EHR from tabular-transformer by $\mathbf{h}_{tab}=\mathcal{T}_{tab}(\mathbf{x}_{t\leq \tau})$ and textual transformer by $\mathbf{h}_{txt}=\mathcal{T}_{txt}(g(\mathbf{x}_{t\leq \tau}))$, we utilize a late-fusion strategy~\cite{shi2021multimodal,lu2023mug} to calculate the final probability of the pediatric patient's cardiac arrest onset. 
We still use the powerful Transformer architecture as the major technical component, with a concatenation operation to pool the two representations
\begin{equation}
\hat{p}=sigmoid(W(\mathcal{F}_{1:M}(\mathbf{h_{tab}},\mathbf{h_{txt}}))+b),
\end{equation}
where $\hat{p} \in [0, 1]$ denotes the predicted probablity of CA onset, $\mathcal{F}_{1:M}(\cdot)$ denotes the multiple transformer layers, and $\mathcal{W}(\cdot)+b$ denotes the last prediction layer.  

The focal loss~\cite{ross2017focal} is used to train the three modules of \modelname to deal with the class imbalance (CA incidence=3.1\%)
\begin{equation}
FL(p_{true})=-(1-p_{true})^{\gamma}\log(p_{true}),
\end{equation}
where
\begin{equation}
p_{true}=
\left\{
    \begin{array}{ c l }
    \hat{p}, & \quad \textrm{if } y=1,\\
    1-\hat{p}, & \quad \textrm{otherwise},\\
    \end{array}
\right.
\end{equation}
and $\gamma$ is a hyperparameter focusing on reduce the loss contribution from well-classified examples ($p_{true} \approx1$).

% a late-fusion model
% where separate neural operations are conducted
% on each data type and extracted high-level representations are aggregated near the output layer.
% Specifically, MLPs are used for tabular modality, and transformers are used for text and image
% modalities. After that, dense vector embeddings
% from the last layer of each network are pooled
% into one vector, and the final prediction is obtained via an additional two-layer MLP
% concat is used as default.

\section{Experimental Results}

\subsection{Experimental Settings}
We use our curated cardiac arrest cohort from CHOA-CICU to conduct 5-fold cross-validation. We further ensure that no patient appears in both the training and testing sets for an unbiased evaluation.
As defined in Sec.~\ref{sec:study}, our research problem, early risk prediction of CA, is formulated asa a binary classification task. Accordingly, we employ a broad set of binary classification metrics, including balanced accuracy (Bal\_Acc), F1 score, Matthews Correlation Coefficient (MCC), Area Under the Precision-Recall Curve (AUPRC), and Area Under the Receiver Operating Characteristic Curve (AUROC), to comprehensively assess model performance while accounting for the inherent class imbalance (admission-wise CA incidence $=3.1\%$). 
% talk about the metrics a bit
In our evaluation, CA events are treated as the positive class. 
Among these metrics, Bal\_Acc is a specifically designed metric to avoid inflated performance on imbalanced dataset, which is essentially arithmetic mean of sensitivity and specificity in binary classification case. F1 is a harmonic mean of the precision and recall. 
MCC is defined as
\begin{equation}
    \text{MCC}=\frac{tp\times tn - fp \times fn}{\sqrt{(tp+fp)(tp+fn)(tn+fp)(tn+fn)}},
\end{equation}
where is generally regarded as a balanced metric for imbalanced case. A MCC of $+1$ presents a perfect prediction, $0$ an random prediction, and $-1$ an inverse prediction. AUROC and AUPRC both assess trade-offs between key performance rates. While AUROC is widely used—providing a threshold-independent evaluation of the balance between TPR and FPR and serving as a valuable benchmark for comparison with prior work—AUPRC is often more informative in scenarios with class imbalance~\cite{tomavsev2021use}, as it emphasizes the trade-off between precision and recall and focuses on performance for the minority (positive) class.

\begin{table*}[ht!]
\centering
\centering
\caption{Experimental results on CHOA-CICU curated cohort (numbers in percentage).}
\label{tab:CHOA_main_exp}
\begin{tabular}{c|ccccc}
\toprule
Model & Bal\_Acc & F1 & MCC & AUPRC &AUROC\\
\multicolumn{6}{c}{\textit{Clinician Derived Risk Scores}} \\
KNN-Dtw &  $51.65 \pm 3.71$ & $2.06 \pm 4.61$ & $1.36 \pm 3.35$ & $3.81 \pm 0.91$ & $55.27 \pm 4.21$ \\
SVC-Gak &  $50.00 \pm 0.00$ & $0.00 \pm 0.00$ & $0.00 \pm 0.00$ & $6.19 \pm 2.21$ & $54.72 \pm 3.80$\\
\midrule
\multicolumn{6}{c}{\textit{Last Observed Risk Factors}} \\
KNN-Unif &  $50.22 \pm 0.72$ & $1.18 \pm 2.63$ & $1.83 \pm 5.62$ & $4.53 \pm 1.24$ & $57.41 \pm 2.99$\\
LR &  $49.80 \pm 0.77$ & $2.70 \pm 1.56$ & $-0.39 \pm 1.52$ & $4.60 \pm 0.49$ & $62.92 \pm 1.92$\\
RF-Gini &  $50.00 \pm 0.00$ & $0.00 \pm 0.00$ & $0.00 \pm 0.00$ & $7.05 \pm 0.81$ & $73.29 \pm 2.47$ \\
XGBoost & $\underline{61.03} \pm 3.18$ & $\underline{11.80} \pm 1.96$ & $\underline{10.18} \pm 2.45$ & $6.88 \pm 1.18$ & $72.35 \pm 4.00$ \\
LightGBM &  $53.74 \pm 5.28$ & $8.62 \pm 5.92$ & $5.24 \pm 6.96$ & $5.82 \pm 2.26$ & $61.42 \pm 11.06$\\
TabNN &  $50.84 \pm 0.64$ & $3.90 \pm 2.33$ & $1.82 \pm 1.68$ & $5.34 \pm 0.79$ & $62.16 \pm 3.64$\\
\hline
\multicolumn{6}{c}{\textit{Numerical Time-Series Risk Factors}} \\
TResNet & $53.93 \pm 2.80$ & $10.32 \pm 4.54$ & $7.47 \pm 4.83$ & $7.53 \pm 1.95$ & $72.21 \pm 7.22$\\
gMLP & $52.51 \pm 1.64$ & $8.05 \pm 3.54$ & $5.11 \pm 3.54$ & $\underline{8.72} \pm 3.13$ & $\textbf{75.04} \pm 3.52$ \\
\hline
\multicolumn{6}{c}{\textbf{\textit{Our Proposed Tabular-Textual Multimodal Risk Factors}}} \\
\modelname & $\mathbf{63.08}\pm 3.65$ & $\mathbf{14.31}\pm 2.28$ & $\mathbf{13.18}\pm 2.55$ & $\textbf{9.15} \pm 3.16$ & $\underline{73.99} \pm 3.16$\\
\bottomrule
\end{tabular}
\vspace{-0.3cm}
\end{table*}

\subsection{Compared Baseline Models}
To comprehensively evaluate the effectiveness of our proposed model, we include ten other AI models. As we summarize before that the EHR data used for pediatric CA is heterogeneous and multi-resolution, the ten AI models can be categories into the following three groups with different EHR feature processing techniques.
\subsubsection{Clinician Derived Risk Scores} We use Pediatric Early Warning Score (PEWS)~\cite{monaghan2005pews} to process EHR data, which involves assessing a range of vital signs and clinical observations—such as heart rate, respiratory rate, blood pressure, oxygen saturation, and behavioral changes—assigning scores based on deviations from normal values, and summing these to produce an overall risk score. The PEWS scores are recorded at various time points by professionals in the EHR, resulting in univariate, variable-length time-series data for each patient.
We then employ K-nearest neighbors with dynamic time wrapping~\cite{sakoe1978dtw} (KNN-Dtw) and support vector classifier with global alignment kernel~\cite{cuturi2011gak} (SVC-Gak). Both dynamic time warping and the global alignment kernel are specifically designed to capture nonlinear similarities in univariate time series of unequal lengths. We implement KNN-Dtw and SVC-Gak based on \textit{tslearn} codebase~\cite{tslearn}.
\subsubsection{Last Observed Risk Factors} We implement a dedicated aggregation process, $LAST(\cdot)$, to transform EHR data into a multivariate static tabular format by capturing the most recent observations of risk factors.  This transformation enables the use of a wide range of classical tabular AI models, including KNN with uniform weights (KNN-Unif), logistic regression (LR), random forest with Gini impurity (RF-Gini), extreme gradient boosting (XGBoost)~\cite{chen2016xgboost}, light gradient-boosting machine (LightGBM)~\cite{ke2017lightgbm}, and tabular neural network (tabNN)~\cite{ke2018tabnn}. In this process, numerical features remain unchanged, categorical features are converted into numerical values via ordinal encoding, and missing values are not imputed but are instead assigned a designated NaN value. All tabular models are implemented based on \textit{AutoGluon} codebase~\cite{erickson2020autogluon}.
\subsubsection{Numerical Time-Series Risk Factors} We perform a resolution unification process on the EHR data by discretizing the time axis into 1-hour intervals. For each risk factor, repeated recorded values within a 1-hour bucket are aggregated using the $mean(\cdot)$ operator. As a result, we obtain 82 distinct numerical time-series features with a fixed length of 24, representing data from the first 24 hours after admission. Categorical risk factors are excluded from this process, as there is no well-defined aggregation operator for them. To capture both the interactions among multivariate time-series features and their long-term dependencies, we employ state-of-the-art AI models. In particular, we leverage TResNet~\cite{wang2017TResNet}, which utilizes residual convolutional neural networks, and gMLP~\cite{liu2021gMLP}, which is based on a gated multilayer perception architectures. Both deep neural networks are trained with focal loss~\cite{ross2017focal} and are further calibrated with Top-K percentile based best threshold to deal with class imbalance. All time-series deep neural networks are implemented based on \textit{tsai} codebase~\cite{tsai}.

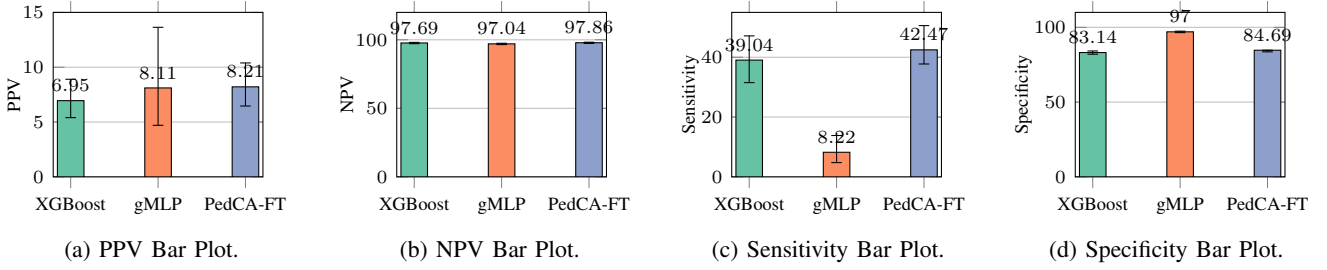
\begin{figure*}[ht!]
  \begin{subfigure}[b]{0.245\textwidth}
  \centering
  \begin{tikzpicture}
    \begin{axis}[
        width=\linewidth,
        ylabel style={font=\scriptsize,yshift=-0.6em},
        y tick label style={font=\scriptsize},
        x tick label style={font=\scriptsize},
        ybar,
        ymajorgrids,
        symbolic x coords={XGBoost, gMLP, PedCA-FT},
        ylabel={PPV},
        ymin=0,
        ymax=15,
        bar shift=0pt,
        nodes near coords, 
        nodes near coords style={font=\scriptsize}, 
    ]
        \addplot[
            fill=Set2-A,
            ybar,
            error bars/.cd,
            y dir=both,
            y explicit,
        ] coordinates {
            (XGBoost, 6.95) += (0, 1.95) -= (0, 1.55)
        };
        \addplot[
            fill=Set2-B,
            ybar,
            error bars/.cd,
            y dir=both,
            y explicit,
        ] coordinates {
            (gMLP, 8.11) += (0, 5.53) -= (0, 3.41)
        };
        \addplot[
            fill=Set2-C,
            ybar,
            error bars/.cd,
            y dir=both,
            y explicit,
        ] coordinates {
            (PedCA-FT, 8.21) += (0, 2.18) -= (0, 1.75)
        };
    \end{axis}
\end{tikzpicture}
  \caption{PPV Bar Plot.}
  \label{fig:CHOA-PPV}
  \end{subfigure}
  % sfig2
  \hfill
  \begin{subfigure}[b]{0.245\textwidth}
  \centering
  \begin{tikzpicture}
    \begin{axis}[
        width=\linewidth,
        ylabel style={font=\scriptsize,yshift=-0.6em},
        y tick label style={font=\scriptsize},
        x tick label style={font=\scriptsize},
        ybar,
        ymajorgrids,
        symbolic x coords={XGBoost, gMLP, PedCA-FT},
        ylabel={NPV},
        ymin=0,
        ymax=120,
        bar shift=0pt,
        nodes near coords, 
        nodes near coords style={font=\scriptsize}, 
    ]
        \addplot[
            fill=Set2-A,
            ybar,
            error bars/.cd,
            y dir=both,
            y explicit,
        ] coordinates {
            (XGBoost, 97.69) += (0, 0.43) -= (0, 0.52)
        };
        \addplot[
            fill=Set2-B,
            ybar,
            error bars/.cd,
            y dir=both,
            y explicit,
        ] coordinates {
            (gMLP, 97.04) += (0, 0.45) -= (0, 0.54)
        };
        \addplot[
            fill=Set2-C,
            ybar,
            error bars/.cd,
            y dir=both,
            y explicit,
        ] coordinates {
            ({PedCA-FT}, 97.86) += (0, 0.4) -= (0, 0.51)
        };
    \end{axis}
\end{tikzpicture}
  \caption{NPV Bar Plot.}
  \label{fig:CHOA-NPV}
  \end{subfigure}
  \hfill
  \begin{subfigure}[b]{0.245\textwidth}
  \centering
  \begin{tikzpicture}
    \begin{axis}[
        width=\linewidth,
        ylabel style={font=\scriptsize,yshift=-0.6em},
        y tick label style={font=\scriptsize},
        x tick label style={font=\scriptsize},
        ybar,
        ymajorgrids,
        symbolic x coords={XGBoost, gMLP, PedCA-FT},
        ylabel={Sensitivity},
        ymin=0,
        ymax=55,
        bar shift=0pt,
        nodes near coords, 
        nodes near coords style={font=\scriptsize}, 
    ]
        \addplot[
            fill=Set2-A,
            ybar,
            error bars/.cd,
            y dir=both,
            y explicit,
        ] coordinates {
            (XGBoost, 39.04) += (0, 8.1) -= (0, 7.53)
        };
        \addplot[
            fill=Set2-B,
            ybar,
            error bars/.cd,
            y dir=both,
            y explicit,
        ] coordinates {
            (gMLP, 8.22) += (0, 5.6) -= (0, 3.46)
        };
        \addplot[
            fill=Set2-C,
            ybar,
            error bars/.cd,
            y dir=both,
            y explicit,
        ] coordinates {
            (PedCA-FT, 42.47) += (0, 8.11) -= (0, 4.73)
        };
    \end{axis}
\end{tikzpicture}
  \caption{Sensitivity Bar Plot.}
  \label{fig:CHOA-sensitivity}
  \end{subfigure}
  \hfill
  \begin{subfigure}[b]{0.245\textwidth}
  \centering
  \begin{tikzpicture}
    \begin{axis}[
        width=\linewidth,
        ylabel style={font=\scriptsize,yshift=-0.6em},
        y tick label style={font=\scriptsize},
        x tick label style={font=\scriptsize},
        ybar,
        ymajorgrids,
        symbolic x coords={XGBoost, gMLP, PedCA-FT},
        ylabel={Specificity},
        ymin=0,
        ymax=110,
        bar shift=0pt,
        nodes near coords, 
        nodes near coords style={font=\scriptsize}, 
    ]
        \addplot[
            fill=Set2-A,
            ybar,
            error bars/.cd,
            y dir=both,
            y explicit,
        ] coordinates {
            (XGBoost, 83.14) += (0, 1.06) -= (0, 1.12)
        };
        \addplot[
            fill=Set2-B,
            ybar,
            error bars/.cd,
            y dir=both,
            y explicit,
        ] coordinates {
            (gMLP, 97.0) += (0, 0.45) -= (0, 0.54)
        };
        \addplot[
            fill=Set2-C,
            ybar,
            error bars/.cd,
            y dir=both,
            y explicit,
        ] coordinates {
            (PedCA-FT, 84.69) += (0, 0.02) -= (0, 1.08)
        };
    \end{axis}
\end{tikzpicture}
  \caption{Specificity Bar Plot.}
  \label{fig:CHOA-specificity}
  \end{subfigure}
  \caption{Bar plots with error bars illustrating PPV, NPV, Sensitivity, and Specificity.}
  \label{fig:bar_chart}
\end{figure*}

\begin{figure*}
\centering
\begin{tikzpicture}
\begin{semilogyaxis} [
width=0.95\linewidth,
height=0.3\linewidth,
%log ticks with fixed point,
%Capillary refill time (CRT) 
% MEAN PLATELET VOL (MPV)
% Race (first Race)
symbolic x coords={LLE CRT,Bilirubin,Hematocrit,MPV,LLE Temp,CO\textsubscript{2},Rhythm Strip,Sex,Resp Effort,Anion Gap,Race,FIO\textsubscript{2},L Breath Sounds, Blender O\textsubscript{2},D5 1/2NS},
xtick=data,
ymode=log, % Set y-axis to logarithmic scale
log basis y={10}, % Base 10 logarithm for the y-axis
xticklabel style = {font=\scriptsize, rotate=-15},
]
\addplot+[only marks] plot[error bars/.cd, y dir=both, y explicit]
coordinates{
    (LLE CRT,0.00986033870379835) +- (0.0147661644336152,0.00495451297398151)
    (Bilirubin,0.00889949693092115) +- (0.0126273414402402,0.00517165242160214) 
    (Hematocrit,0.00771484551610147) +- (0.0106733043693973,0.00475638666280566)
    (MPV,0.00722791639121327) +- (0.0155413424057195,-0.00108550962329301)
    (LLE Temp,0.00654834543779) +- (0.0117898135027299,0.00130687737285006)
    (CO\textsubscript{2},0.00643982552509323) +- (0.0117188530496206,0.00116079800056584)
    (Rhythm Strip,0.0063570944708815) +- (0.0135076080519946,-0.00079341911023163)
    (Sex,0.0060215957123226) +- (0.007101505027428,0.00494168639721721)
    (Resp Effort,0.00529990635784179) +- (0.00995787056878017,0.000641942146903415)
    (Anion Gap,0.00505141865792101) +- (0.00971814564992059,0.000384691665921432)
    (Race,0.00492530982150281) +- (0.00893863381241345,0.000911985830592171)
    (FIO\textsubscript{2},0.00478955726231307) +- (0.00737036239483328,0.00220875212979286)
    (L Breath Sounds,0.00462497034406318) +- (0.00745013982514881,0.00179980086297755)
    (Blender O\textsubscript{2},0.00454991688743892) +- (0.00556981498211942,0.00353001879275842)
    (D5 1/2NS,0.00447869625451364) +- (0.00641773512494759,0.00253965738407969)
};
\end{semilogyaxis} 
\end{tikzpicture}
\caption{Feature importance analysis using feature permutation. Error bar indicates p95 high and p95 low.}
\label{fig:feat_important}
\vspace{-0.4cm}
\end{figure*}
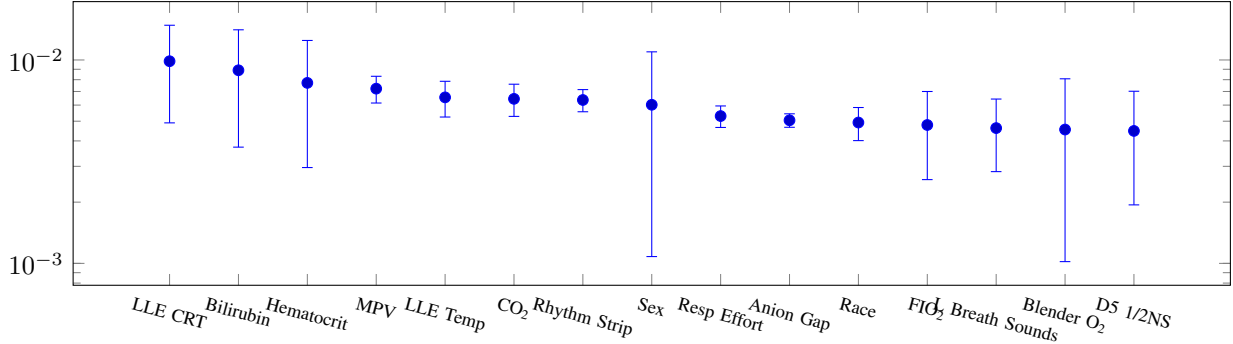

\subsection{Experimental Results \& Discussion}
Table~\ref{tab:CHOA_main_exp} summarizes the performance of our proposed \modelname to ten baseline models on various evaluation metrics. 
Notably, \modelname achieves the highest Bal\_acc, F1, MCC, and AUPRC outperforming all baseline models by an average improvement of $10.71$, $9.45$, $9.55$, and $3.10$ separately. 
Additionally, \modelname ranks second in AUROC, with an average improvement of $11.57$ except for gMLP model. 
Among baseline models, XGBoost achieves the runner-up in Bal\_Acc, F1, and MCC, while gMLP achieves the second best in AUPRC and the highest AUROC. 
The slightly lower AUROC of \modelname compared to gMLP can be attributed to AUROC's emphasis on overall ranking separability, which does not necessarily translate into optimal classification decisions~\cite{tomavsev2021use}, particularly in highly imbalanced datasets. AUROC measures the ability of a model to distinguish between control and case classes across all possible classification thresholds. However, it averages performance across all thresholds and gives equal importance to the classification of both the majority (control) and minority (CA) classes. In highly imbalanced datasets such as the pediatric CA prediction scenario, this can be problematic because the model might perform well on predicting non-CA patients and still achieve a high AUROC, even if it struggles to correctly identify patients at high risk of CA. 
Furthermore, our results highlight the effectiveness of multimodal data fusion. Models based on tabular representations (i.e., those using last observed risk factors) generally achieve superior Bal\_Acc, F1, and MCC, whereas models leveraging time-series representations tend to perform better in AUPRC and AUROC. Integrating a tabular transformer to capture complex interactions among high-dimensional tabular risk factors and a pre-trained textual transformer to effectively model the dynamics of textualized time-series risk factors, \modelname delivers SOTA performance for CA risk prediction.

% bar chart
We further analyze XGBoost, gMLP, and \modelname in terms of Positive Predictive Value (PPV), Negative Predictive Value (NPV), Sensitivity, and Specificity, as shown in Figure~\ref{fig:bar_chart}, to align with clinical priorities. Overall, \modelname achieves the most balanced performance across all four metrics.
Specifically, \modelname demonstrates a slight advantage in both PPV and NPV, while gMLP exhibits high variance in PPV, as indicated by the wide error bars. Notably, \modelname attains the highest sensitivity, which is particularly crucial for screening CA, where missing cases can have severe consequences. At the same time, \modelname maintains competitive specificity, striking a careful balance between sensitivity and specificity—two metrics that often trade off inversely.
These observations further validate the limitations of AUROC as a sole evaluation metric, highlighting its tendency to be overly optimistic or misleading in highly imbalanced settings. While AUROC emphasizes ranking separability, it does not directly reflect clinical utility, whereas metrics like sensitivity and PPV better capture a model’s practical impact on early CA detection.

\subsection{Feature Importance Analysis}
We use permutation importance method~\cite{altmann2010permutation} to calculate the feature importance scores for our proposed perturbed copy of the data where this feature's values have been randomly shuffled across rows. We set number of different permutation shuffles as $5$, and the confidence level as $0.95$. Top 15 important features with $p\_value \leq 0.05$ are presented in Fig.~\ref{fig:feat_important}.
Our analysis reveals that key clinical indicators, such as capillary refill (CRT), total bilirubin level, and hematocrit, are among the most important predictors for pediatric cardiac arrest. 
Notably, factors related to hemodynamic stability and oxygenation/ventilation—including CO\textsubscript{2} level on arterial blood gas, and FIO\textsubscript{2},—also rank highly, emphasizing their critical role in patient monitoring. Furthermore, the inclusion of demographic factors such as sex highlights the multifactorial nature of cardiac arrest risk in the pediatric population.
Overall, this feature importance analysis not only identifies clinical risk factors but also uncovers novel predictors that may enhance early risk detection and guide future targeted interventions.

%\subsection{Ablation Study of Two Base Models}
\section{Conclusion}
Electronic health records contain a wealth of patient information that can be leveraged to study risk factors for the early detection of pediatric cardiac arrest. 
In this paper, we introduce \modelname, a novel multimodal fused transformer that effectively handle the heterogeneous multi-resolution EHR data. Extensive experiments on a curated pediatric cohort demonstrate the effectiveness of \modelname, indicating its potential translational impact in clinical settings.
In the future, we plan to (1) extend our methodology to more diverse datasets and incorporate other modalities (\textit{e.g.}, clinical notes, waveforms) to further enhance predictive accuracy; (2) Explore continuous risk monitoring scenarios that enable real-time prediction updates at fixed time intervals or when new risk factor data becomes available.



\section*{ACKNOWLEDGMENT}

This work was supported in part by the Nell Hodgson
Woodruff School of Nursing (NHWSN) Center for Data Science of Emory University, the NHWSN High Performance Computing cluster and the NHWSN IT Department.

%%%%%%%%%%%%%%%%%%%%%%%%%%%%%%%%%%%%%%%%%%%%%%%%%%%%%%%%%%%%%%%%%%%%%%%%%%%%%%%%

\bibliographystyle{IEEEtran}
\bibliography{mybib}

\end{document}